\documentclass{article}
\PassOptionsToPackage{numbers, compress}{natbib}
\usepackage[preprint]{neurips_2023}
\usepackage[utf8]{inputenc} 
\usepackage[T1]{fontenc}    
\usepackage{hyperref}       
\usepackage{url}            
\usepackage{booktabs}       
\usepackage{amsfonts}       
\usepackage{nicefrac}       
\usepackage{microtype}      
\usepackage{xcolor}         
\usepackage{amsmath,amssymb}
\usepackage{algorithm}
\usepackage{algpseudocode}
\usepackage{listings}
\usepackage{graphicx}
\usepackage{booktabs}
\usepackage{bm}
\usepackage{subfig}
\usepackage{multirow}

\title{DFormer: Diffusion-guided Transformer for Universal Image Segmentation}

\author{%
  Hefeng Wang$^1$, Jiale Cao$^1$, Rao Muhammad Anwer$^2$, Jin Xie$^3$,  \\
  \textbf{Fahad Shahbaz Khan}$^2$, 
  \textbf{Yanwei Pang}$^{1,4}$\\
  $^1$ Tianjin University~~$^2$ Mohamed bin Zayed University of Artificial Intelligence\\
  $^3$ Chongqing University~~$^4$ Shanghai Artificial Intelligence Laboratory\\
  \texttt{\{wanghefeng,connor,pyw\}@tju.edu.cn} \\
  \texttt{\{rao.anwer,fahad.khan\}@mbzuai.ac.ae~~xiejin@cqu.edu.cn} \\
}

\begin{document}

\maketitle

\begin{abstract}

This paper introduces an approach, named DFormer, for universal image segmentation. The proposed DFormer views universal image segmentation task as a denoising process using a diffusion model. DFormer first adds various levels of Gaussian noise to ground-truth masks, and then learns a  model to predict denoising masks from  corrupted masks. Specifically, we take deep pixel-level features along with the noisy masks as inputs to generate mask features and attention masks, employing diffusion-based decoder to perform mask prediction gradually. At inference, our DFormer directly predicts the masks and corresponding categories from a set of randomly-generated masks. Extensive experiments reveal the merits of our proposed contributions on different image segmentation tasks: panoptic segmentation, instance segmentation, and semantic segmentation. Our DFormer outperforms the recent diffusion-based panoptic segmentation method Pix2Seq-$\mathcal{D}$ with a gain of 3.6\% on MS COCO \texttt{val2017} set. Further, DFormer achieves promising semantic segmentation performance outperforming the recent diffusion-based method by  2.2\%  on ADE20K \texttt{val} set. Our source code and models will be publicly on \url{https://github.com/cp3wan/DFormer}.

\end{abstract}

\section{Introduction}
Image segmentation  aims to group the pixels within an image into different units. There are different notations about the grouping, such as semantic categories or instances. Therefore, a variety of image segmentation tasks have been proposed, including semantic segmentation \cite{Long_FCN_CVPR_2015}, instance segmentation \cite{He_MaskRCNN_ICCV_R2017}, and panoptic segmentation \cite{Kirillov_PS_CVPR_2019}. Semantic segmentation groups the pixels into different semantic categories, whereas instance segmentation groups the pixels into different instances. On the other hand, panoptic segmentation not only groups the pixels of things into different instances, but also segments the stuffs into different semantic categories, which can be seen as a joint task of semantic segmentation and instance segmentation.

In the past years, researchers have proposed specialized architectures for different image segmentation tasks. For instance, different variants of fully-convolutional networks \cite{Long_FCN_CVPR_2015,deeplabV2,Zhao_PSPNet_CVPR_2017} are used for pixel-level classification in semantic segmentation.  Detect-then-segment \cite{He_MaskRCNN_ICCV_R2017,Chen_HTC_CVPR_2019,Cao_D2Det_CVPR_2020} and end-to-end \cite{Wang_SOLOv2_NIPS_2020,Wu_SeqFormer_ECCV_2022,Wang_LES_NeurIPS_2022} architectures are built for instance segmentation, whereas split-then-merge pipelines \cite{Kirillov_PFPN_CPVR_2019,Li_APS_CPVR_2019} are designed to combine semantic and instance segmentation for panoptic segmentation. While these specialized methods have achieved great success in each individual segmentation task, they typically struggle to effectively generalize to different image segmentation tasks.

To address the aforementioned issue, few existing works \cite{Zhang_KNet_NeurIPS_2021,Cheng_MaskFormer_NeurIPS_2021,Wang_CLUSTSEG_ICML_2023} attempt to develop a universal architecture that  performs different image segmentation tasks through a unified framework. Inspired by the transformer-based object detector, DETR \cite{Carion_DETR_ECCV_2020}, these methods view image segmentation as an end-to-end set prediction problem. For instance, K-Net \cite{Zhang_KNet_NeurIPS_2021} employs  a group of learnable kernels to dynamically segment instance and stuff, while  Mask2Former \cite{Cheng_Mask2Former_CVPR_2022} introduces a masked-attention mask transformer to perform mask classification and mask prediction.  Recently, diffusion model-based methods have also been explored for image segmentation. For instance, Chen \textit{et al.} \cite{Chen_GenD_arXiv_2022} employ bit diffusion for panoptic segmentation, while Ji \textit{et al.} \cite{Ji_DDP_arXiv_2023}  propose to concatenate noise and deep features for semantic segmentation. However, the diffusion model-based methods still lag behind existing universal image segmentation methods. In this work, we investigate the problem of designing an effectively diffusion model-based transformer approach that can achieve competitive universal image segmentation performance.

\begin{figure}%
\centering
\footnotesize
\begin{minipage}[c]{0.4\textwidth}%
\centering
\includegraphics[width=1\textwidth]{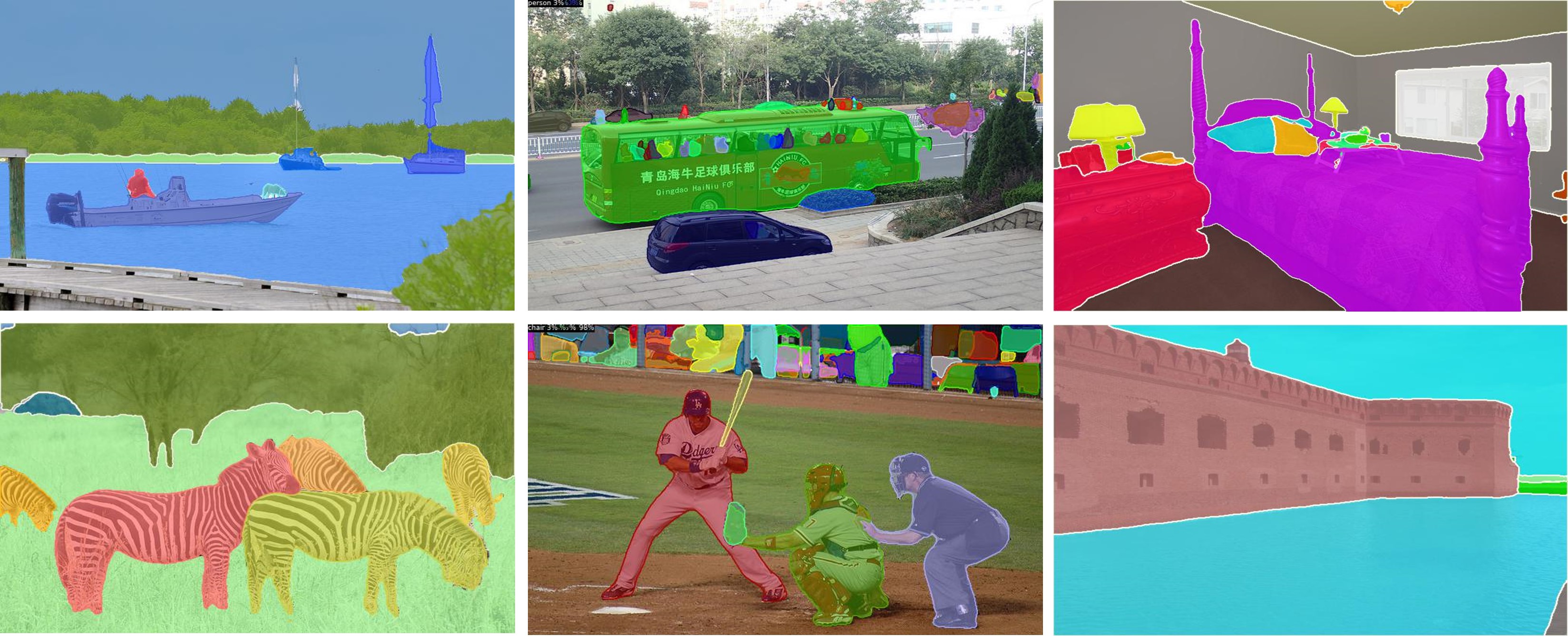}
\end{minipage}\hfill
\begin{minipage}[c]{0.6\textwidth}%
\centering
\footnotesize
\resizebox{0.9\linewidth}{!}{
\begin{tabular}{| r | c | c c  c |}
\hline
Method & Backbone & Pan. & Inst.  & Sem.\\
\hline
\hline

Pix2Seq-$D$ \cite{Chen_GenD_arXiv_2022} & ResNet-50&  47.5  &   -  & -\\
DiffusionInst \cite{Gu_DiffInst_arXiv_2022} & ResNet-50&  -  &  37.3  & -\\
\textbf{DFormer (Ours)} & ResNet-50  & 51.1     &     42.6   &     46.7\\
\hline
DDP \cite{Ji_DDP_arXiv_2023} & Swin-T &  -  &   -  &  46.1\\
\textbf{DFormer (Ours)} & Swin-T &  52.5    &      44.4   &     48.3\\
\hline
\end{tabular}}
\end{minipage}
\caption{\textbf{Left:} Image segmentation results of our DFormer, using the same architecture, on three  different tasks: panoptic segmentation (Column 1), instance segmentation (Column 2) and semantic segmentation (Column 3). 
\textbf{Right:} Comparison with other diffusion model-based segmentation methods on three tasks. `Pan.' indicates PQ of panoptic segmentation on COCO \texttt{val2017}, `Inst.' indicates AP of instance segmentation on COCO \texttt{val2017}, `Sem.' indicates mIoU of semantic segmentation on ADE20K \texttt{val}. Our DFormer achieves consistent improvements on all three tasks.}
\label{fig:vis2tab}
\vspace{-0.5cm}
\end{figure}

We propose a diffusion-guided transformer framework, named DFormer, for universal image segmentation. Our DFormer views image segmentation as a generative task from noisy masks. During training, we  add various level of Gaussian noise to ground-truth masks to obtain noisy masks. Afterwards, we generate the attention mask with binary threshold, and aggregate noisy masks and deep pixel-level features to produce mask features. We then feed them to the transformer decoder to predict the ground-truth masks for each mask feature with masked attention. At inference, we first generate a set of noisy masks and then employ the diffusion-based  decoder to predict  masks and corresponding object categories.

We conduct experiments on three different image segmentation tasks: semantic segmentation, instance segmentation, and panoptic segmentation. Our experimental results show the merits of the proposed contributions achieving promising performance (see Fig. \ref{fig:vis2tab}) on three segmentation tasks with a single architecture. 
With the backbone ResNet50, our DFormer achieves a PQ score of 51.1\% on MS COCO \texttt{val2017} set for panoptic segmentation, thereby outperforming the recent diffusion-based panoptic segmentation method Pix2Seq-$\mathcal{D}$ \cite{Chen_GenD_arXiv_2022} with an absolute gain of 3.6\% (see Fig.\ref{fig:vis2tab}). With the backbone Swin-T, our DFormer performs favorably against the recent diffusion-based semantic segmentation method DDP \cite{Ji_DDP_arXiv_2023} by achieving mIoU score of 48.3\% on ADE20K \texttt{val} set for semantic segmentation.

\section{Related works}
\label{sec:relatedworks}

\noindent\textbf{Semantic segmentation.} Semantic segmentation  aims to group the pixels within an image into different semantic categories (\textit{e.g.,} person, car, and road). With the advent of deep learning, semantic segmentation has achieved excellent progress. FCN \cite{Long_FCN_CVPR_2015} is one of the earliest works for semantic segmentation, which adopts fully convolutional networks for pixel-level classification. Afterwards, the researchers developed different FCN variants for improved semantic segmentation. For instance, some works focus on local contextual information aggregation with encoder-decoder structure \cite{Zhao_PSPNet_CVPR_2017,Cao_TripleNet_CVPR_2019,Wang_HRNet_TPAMI_2019} or spatial pyramid structure  \cite{Chen_DeepLabV3_arXiv_2017,deeplabV2}. Further, several works exploit attention mechanism for non-local context aggregation \cite{Fu_DANet_CVPR_2019,Huang_CCNet_ICCV_2019,Yuan_OCNet_IJCV_2021}, whereas other works aim at real-time design \cite{Yu_BiSeNet_ECCV_2018,Fan_RBiSeNet_CVPR_2021}. Recently, transformer-based methods have been proposed in literature, such as transformer-based network design \cite{Yuan_HRFormer_NeurIPS_2021,Gu_HRViT_CVPR_2022,Xie_SegFormer_NeurIPS_2021} and transformer-based  segmentation head  \cite{Zheng_SETR_CVPR_2021,Strudel_Segmeter_ICCV_2021}.

\noindent\textbf{Instance segmentation.} Instance segmentation classifies and segments different object instances (\textit{e.g.,} person and car), and have been addressed through two-stage, single-stage and end-to-end approaches. Two-stage approaches \cite{He_MaskRCNN_ICCV_R2017,Huang_MSRCNN_CVPR_2019,Kirillov_PointRend_CVPR_2020} are usually built on the two-stage object detectors, that first generate some candidate region proposals and then predict object masks within cropped region proposals. 

In contrast to two-stage approaches, single-stage approaches \cite{Chen_TensorMask_ICCV_2019,Tian_CondInst_ECCV_2020,Cao_SipMask_ECCV_2020} directly perform mask prediction using fully-convolutional networks. 

Recently, with the success of end-to-end object detectors, such as DETR \cite{Carion_DETR_ECCV_2020}, several works have explored end-to-end instance segmentation. QueryInst \cite{Fang_QueryInst_ICCV_2021} designs a dynamic mask head for query-based instance segmentation. SOLQ \cite{Dong_SOLQ_NeurIPS_2021} builds a unified query representation for instance classes, locations, and masks. SOTR \cite{Guo_SOTR_ICCV_2021} develops a twin attention mechanism into transformer for instance segmentation. In addition, FastInst \cite{He_FastInst_CVPR_2023} and SparseInst \cite{Cheng_SparseInst_CVPR_2022} focus on efficient design.

\noindent\textbf{Panoptic segmentation.} Panoptic segmentation \cite{Kirillov_PS_CVPR_2019,Kirillov_PFPN_CPVR_2019} combines instance segmentation and semantic segmentation into a unified task. At first, several works \cite{Xiong_UPSNet_CVPR_2019,Liu_OANet_CVPR_2019,Li_APS_CPVR_2019,Lazarow_OCFusion_CVPR_2020,Li_UTI_CVPR_2020} explored to use different branches for two sub-tasks, and design a fusion module to combine the results of these branches for panoptic segmentation. For instance, Panoptic FPN \cite{Kirillov_PFPN_CPVR_2019} attaches two different head-networks on the shared feature pyramid, and employs a post-processing operation to combine the results of two head-networks. Though these methods have achieved promising results, they are relatively complex. Afterwards, several works \cite{Li_PFCN_CPVR_2021,Wang_MaxDeepLab_CVPR_2021,Li_PSegFormer_CVPR_2022,Yu_KMaX_ECCV_2022} explore unifying thing and stuff segmentation into a single network. For instance, Panoptic FCN \cite{Li_PFCN_CPVR_2021} first generates the kernels for both thing and stuff, and then employs dynamic convolutions for prediction. Panoptic SegFormer \cite{Li_PSegFormer_CVPR_2022} introduces a query decoupling strategy, where thing and stuff predictions are based on different queries.

\noindent\textbf{Universal image segmentation.} Universal image segmentation aims to unify semantic, instance, and panoptic segmentation tasks into a single network. Inspired by the success of vision transformer for object detection \cite{Carion_DETR_ECCV_2020} that treats detection as a set prediction problem,  researchers \cite{Cheng_MaskFormer_NeurIPS_2021,Zhang_KNet_NeurIPS_2021} look into exploring a universal architecture for different image segmentation tasks. K-Net \cite{Zhang_KNet_NeurIPS_2021} learns a set of learnable kernels to predict both instance and semantic categories.  MaskFormer \cite{Cheng_MaskFormer_NeurIPS_2021} and Mask2Former \cite{Cheng_Mask2Former_CVPR_2022} treat image segmentation as  mask classification problem, where they predict  mask embeddings and pixel-level embeddings, and use their dot product for mask prediction.

\noindent\textbf{Diffusion models for perception tasks.} Diffusion models \cite{Ho_DDPM_NeurIPS_2020} aim to recover the sample from noise with a parameterized Markov chain, that has achieved promising results on the image generation task. Recently, some works started to explore diffusion models for perception tasks.  Chen \textit{et al.}  \cite{Chen_DiffusionDet_arXiv_2022} introduce the diffusion model for object detection, where objects are recovered from  noisy boxes with a progressive denoising process.
Baranchuk \textit{et al.} \cite{Baranchuk_Label_ICLR_2022}, Brempong \textit{et al.} \cite{Brempong_DP_CVPRW_2022}, Kim \textit{et al.} \cite{Kim_DARL_ICLR_2023}, Wu \textit{et al.} \cite{Wu_DiffuMask_arXiv_2023} explore label-efficient  segmentation with the aid of diffusion models. Amit \textit{et al.} \cite{Amit_SegDiff_arXiv_2021} propose to sum the deep features and estimated segmentation map for progressive refinement. Wolleb \textit{et al.} \cite{Wolleb_ISE_arXiv_2021} and Ji \textit{et al.} 
 \cite{Ji_DDP_arXiv_2023} propose to use diffusion model for improved segmentation. Chen \textit{et al.}  introduce bit diffusion \cite{Chen_Bit_ICLR_2023} for discrete data generation and extended bit diffusion for panoptic segmentation \cite{Chen_GenD_arXiv_2022}. However, we note that these diffusion models-based methods still lag behind universal image segmentation approaches on different image segmentation tasks. In this work, we set out to bridge this performance gap between diffusion models-based perception methods and universal image segmentation techniques.

\section{Method}
\label{sec:ours}

\subsection{Preliminaries}
\noindent\textbf{Diffusion model.} Diffusion models \cite{Ho_DDPM_NeurIPS_2020} learn a series of state transitions to generate high-quality sample from the noise. During training, a forward diffusion process is first performed by gradually adding Gaussian noise to sample data, and a model is then learned to predict original sample data from noisy  sample data. Specifically, the  diffusion process  generates  noisy sample data $\mathbf{x}_t$ at arbitrary timestep $t$  as
\begin{equation}
q(\mathbf{x}_t|\mathbf{x}_0) = \mathcal{N}(\mathbf{x}_t; \sqrt{\bar\alpha_t}\mathbf{x}_0, (1-\bar\alpha_t)\mathbf{I}),
\label{eq:diff}
\end{equation}
where, the timestep $t$ uniformly ranges from 1 to $T$,  $\bar\alpha_t$ is a monotonically decreasing function from 1 to 0 with the increase of $t$. When the timestep $t$ is equal to $T$, $\bar\alpha_T$ is close to 0, and noisy data $\mathbf{x}_T$ is close to $\mathbf{I}$.  The noise $\mathbf{I}$ is sampled with standard normal distribution. Afterwards, a network model is trained to predict sample data $\mathbf{x}_0$ from $\mathbf{x}_t$ at arbitrary timestep $t$, where the denoising loss can be written as
\begin{equation}
L_{diff} = \frac{1}{2}||f(\mathbf{x}_t,t)-\mathbf{x}_0||^2,
\end{equation}
where, $f$ represents the learned  model.
During inference, the sample data $\mathbf{x}_0$ is gradually recovered from the noisy data $\mathbf{x}_T$ with a series of state transitions $\mathbf{x}_T \rightarrow \mathbf{x}_{T-\Delta} \rightarrow ... \rightarrow \mathbf{x}_{0}$ \cite{Ho_DDPM_NeurIPS_2020,Song_DDIM_ILCR_2021}, where the learned model is repeatedly employed.

\noindent\textbf{Universal image segmentation.}  Image segmentation is viewed as mask classification task, which predicts the  masks and corresponding categories of existing stuffs and things in an image $\mathbf{x}$, that can be written as $\mathcal{M},\mathcal{C}=f(\mathbf{x})$. Recently, the transformer-based image segmentation has achieved promising results, which can be represented as $\mathcal{M},\mathcal{C}=f_{trans}({f_{pixel}(\mathbf{x}),\mathcal{Q}})$, where, $f_{pixel}$ indicates the pixel-level module for pixel-level feature extraction, $f_{trans}$ indicates the transformer module, and $\mathcal{Q}$ indicates the fixed size of learnable queries. In this paper, we aim to re-formulate image segmentation as a denoising process via diffusion model. Instead of using  a set of learnable queries, we directly segment image from noisy masks as $\mathcal{M},\mathcal{C}=f_{d}({f_{pixel}(\mathbf{x}),\mathbf{I}})$, where $\mathbf{I}$ represents noise data, and $f_{d}$ represents diffusion-based decoder.
\begin{figure}[!t]
\centering
\includegraphics[width=0.88\linewidth]{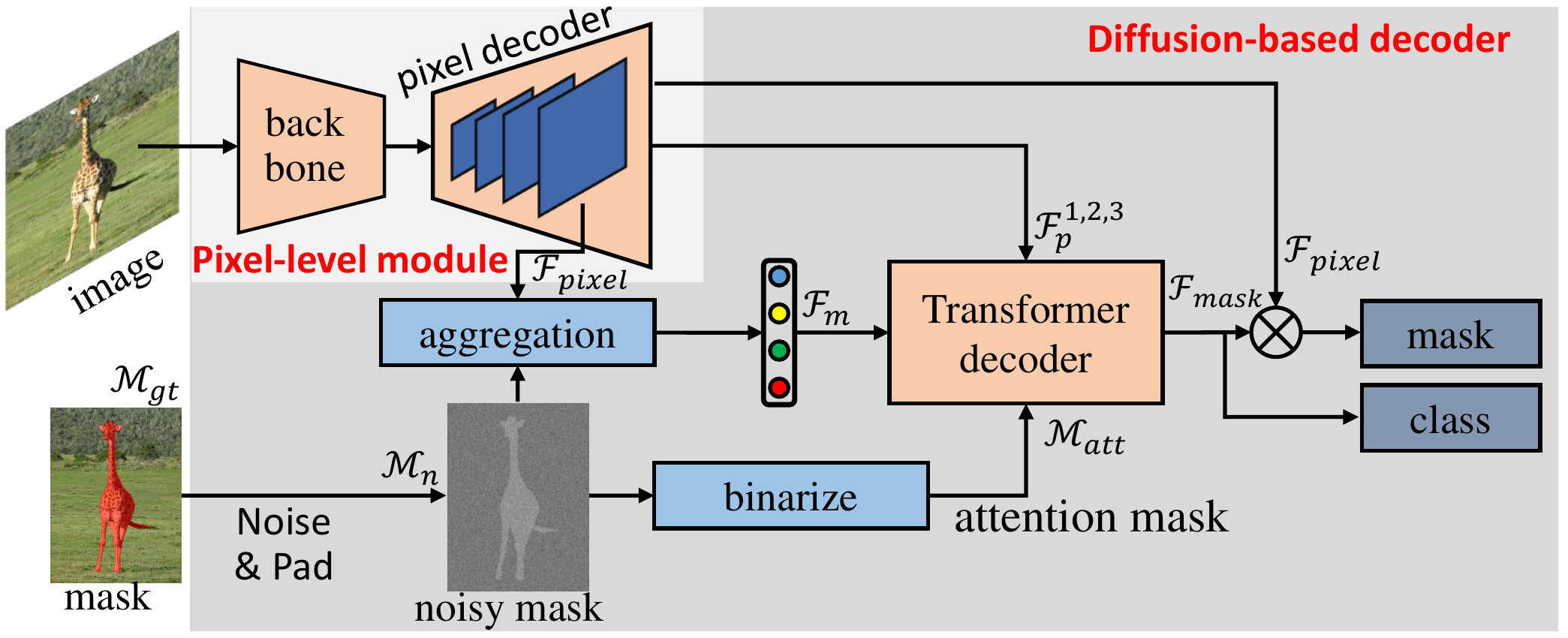}
\caption{\textbf{Overall architecture of our DFormer for universal image segmentation}. We employ pixel-level module, including backbone and pixel decoder, to extract multi-scale features. During training, we perform the diffusion process by adding Gaussian noise to ground-truth mask, and employ a transformer decoder to predict mask and corresponding category from the noisy mask.}
\label{fig:arch}
\vspace{-0.5cm}
\end{figure}
\subsection{Architecture}

Fig. \ref{fig:arch} shows the overall architecture of our diffusion-guided transformer, named DFormer, which  incorporates diffusion model for universal image segmentation. Following \cite{Cheng_Mask2Former_CVPR_2022}, our DFormer comprises a pixel-level module and replaces the standard decoder with the proposed diffusion-based  decoder. The pixel-level module generates  pyramid features $\mathcal{F}_p^i, i=1,2,3$ and pixel-level embeddings $\mathcal{F}_{pixel}$, while our diffusion-based decoder takes noisy masks $\mathcal{M}_n$, pyramid features $\mathcal{F}_p^i, i=1,2,3$, and pixel-level embeddings $\mathcal{F}_{pixel}$ as inputs and predicts mask  embeddings $\mathcal{F}_{mask}$. Finally, we generate the masks by dot product of pixel-level embeddings $\mathcal{F}_{pixel}$  and mask  embeddings $\mathcal{F}_{mask}$, and predict mask categories by feeding mask embeddings to an MLP layer.

\noindent\textbf{Pixel-level module.} Given an input image $\mathbf{x} \in \mathbb{R}^{H\times W\times 3}$, the pixel-level module extracts pyramid features $\mathcal{F}_p^i\in \mathbb{R}^{\frac{H}{S_{i}}\times \frac{W}{S_{i}}\times C},i=1,2,3,4$, where $S_i$  represents the stride of feature map to input image, and $C$ is the number of feature channels. Pixel-level module contains a backbone and a pixel decoder.  Specifically, we first employ the backbone, such as ResNet \cite{He_ResNet_CVPR_2016} or Swin Transformer \cite{Liu_Swin_ICCV_2021}, to extract deep feature of low-resolution $\mathcal{F}_{low}\in \mathbb{R}^{\frac{H}{64}\times \frac{W}{64}\times C}$. Afterwards, we use a pixel decoder to progressively upsample deep feature to generate pyramid features of various resolutions $\mathcal{F}_p^i,i=1,2,3,4$. The first three successive pyramid features $\mathcal{F}_p^i,i=1,2,3$ are fed to successive transformer  decoder layers to generate mask embeddings, and the last pyramid feature $\mathcal{F}_{p}^4$ is used as pixel-level embeddings $\mathcal{F}_{pixel}$.

\noindent\textbf{Diffusion-based decoder.} Based on the generated pyramid features $\mathcal{F}_p^i,i=1,2,3$ and noisy masks $\mathcal{M}_n$, our diffusion-based decoder aims to predict mask embeddings $\mathcal{F}_{mask}$. The diffusion-based decoder stacks $L$ transformer decoder layers as in \cite{Cheng_Mask2Former_CVPR_2022}, where each decoder layer contains a masked-attention layer, a self-attention layer, and a FFN layer. The inputs to each decoder layer are attention masks $\mathcal{M}_{att}$, one of pyramid features $\mathcal{F}_{p}^i$, and mask features $\mathcal{F}_m$. For the first decoder layer,  the attention  masks $\mathcal{M}_{att}$ are generated by thresholding  noisy masks as
$\mathcal{M}_{att} = \mathcal{M}_n > 0$, and mask features $\mathcal{F}_m$ are generated by aggregating pixel-level embeddings $\mathcal{F}_{pixel}$ using noisy masks $\mathcal{M}_{n}$ as weights, which can be written as
\begin{equation}
\mathcal{F}_m = f_{avg}(\mathcal{F}_{pixel} \times f_{norm}(\mathcal{M}_n)),
\end{equation}
where, $f_{avg}$ represents the global averaged pooling operation, and $f_{norm}$ represents the normalization operation.  Except the first decoder layer, the output mask embeddings $\mathcal{F}_{mask}$ of current decoder layer  are used  as input mask features $\mathcal{F}_{m}$ of next decoder layer, and the predicted masks of current decoder layer are used as input attention masks $\mathcal{M}_{att}$ for next decoder layer. The pyramid features $\mathcal{F}_p^i,i=1,2,3$ are alternately used as input image features of successive decoder layers for efficient training. With output mask embeddings $\mathcal{F}_{mask}$, we employ an MLP layer to predict mask classification scores, and employ dot product with pixel-level embeddings to generate predicted masks.

\algrenewcommand\algorithmicindent{0.5em}%
\begin{figure}[t]
\begin{minipage}[t]{0.495\textwidth}
\begin{algorithm}[H]
\small
\caption{\small DFormer training}
\label{alg:train}
\definecolor{codeblue}{rgb}{0.5,0.5,0.5}
\definecolor{codekw}{rgb}{0.85, 0.18, 0.50}
\lstset{
  backgroundcolor=\color{white},
  basicstyle=\fontsize{7.5pt}{7.5pt}\ttfamily\selectfont,
  columns=fullflexible,
  breaklines=true,
  captionpos=b,
  commentstyle=\fontsize{7.5pt}{7.5pt}\color{codeblue},
  keywordstyle=\fontsize{7.5pt}{7.5pt}\color{codekw},
  escapechar={|}, 
}
\begin{lstlisting}[language=python]
def train_loss(images, masks_gt):
  """
  images: [B, 3, H, W]
  masks_gt: [B, *, H, W]
  N: number of masks
  """
  
  # Encode pixel-level pyramid features
  feats = pixel_module(images)
  
  # Pad gt_masks to N
  m_encpad = encode_pad(masks_gt) # [B, N, H, W]
  m_encpad = (m_encpad * 2 - 1) * scale

  # Corrupt gt_masks at various timesteps
  t = randint(0, T) # timestep
  eps = normal(mean=0, std=1)  # noise with the same shape as pd_masks
  m_crpt = sqrt(alpha_cumprod(t)) * m_encpad + 
               sqrt(1 - alpha_cumprod(t)) * eps

  # Predict and compute loss
  m_pred = diff_decoder(m_crpt, feats, t)
  loss = prediction_loss(m_pred, masks_gt)
  
  return loss
\end{lstlisting}
\end{algorithm}
\end{minipage}
\hfill
\begin{minipage}[t]{0.495\textwidth}
\begin{algorithm}[H]
\small
\caption{\small DFormer inference}
\label{alg:sample}
\definecolor{codeblue}{rgb}{0.5,0.5,0.5}
\definecolor{codekw}{rgb}{0.85, 0.18, 0.50}
\lstset{
  backgroundcolor=\color{white},
  basicstyle=\fontsize{7.5pt}{7.5pt}\ttfamily\selectfont,
  columns=fullflexible,
  breaklines=true,
  captionpos=b,
  commentstyle=\fontsize{7.5pt}{7.5pt}\color{codeblue},
  keywordstyle=\fontsize{7.5pt}{7.5pt}\color{codekw},
  escapechar={|}, 
}
\begin{lstlisting}[language=python]
def infer(images, steps):
  """
  images: [B, 3, H, W]
  steps: number of sample steps
  """

  # Extract pixel-level pyramid features
  feats = pixel_module(images)

  # Generate noisy masks
  m_t = normal(mean=0, std=1)   # same shape as m_pred
  
  # uniform sample step size
  times = reversed(linespace(-1, T, steps))

  # [(T-1, T-2), ..., (0, -1)]
  time_pairs = list(zip(times[:-1], times[1:])
  
  for t_now, t_next in range(time_pairs):
    # Predict m_pred from m_t
    m_pred = diff_decoder(m_t, feats, t_now)
        
    # Estimate m_t at time t_next
    m_t = ddim_step(m_t, m_pred, t_now, t_next)
    
  return masks_pred
\end{lstlisting}
\end{algorithm}
\end{minipage}
\vspace{-1em}
\end{figure}

\subsection{Training and Inference}
During training, we perform the diffusion process to generate noisy masks from ground-truth  masks, and train the model to recover ground-truth masks from noisy masks. During inference, we randomly generate noisy masks, and predict  masks and categories with learned models. 

\subsubsection{Training}
Algorithm \ref{alg:train} presents the training details of our DFormer, including pixel-level module, mask  padding and encoding, mask corruption, diffusion-based decoder, and losses. 

\noindent\textbf{Mask  padding and encoding.} We first perform mask padding since the  numbers of objects in different images vary. We pad ground-truth masks $\mathcal{M}_{gt}$ to a fixed number of padded masks $\mathcal{M}_{pad}$. Specifically, we generate some random binary masks $\mathcal{M}_{rand}$ and concatenate it with original ground-truth masks as $\mathcal{M}_{pad}=cat(\mathcal{M}_{gt},\mathcal{M}_{rand})$, where $\mathcal{M}_{pad}\in \mathbb{R}^{N\times H \times W}$. 

Afterwards, we explore two different strategies to encode padded masks before adding noise: binary and random shuffling-based encoding strategy. The binary strategy directly encodes padded masks as binary inputs (0 or 1), while the random shuffling-based strategy encodes the padded mask from 0 to 1 where the object mask pixel randomly samples from 0.5 to 1, and the non-object  pixel randomly samples from 0 to 0.5. 

\noindent\textbf{Mask corruption.} Mask corruption aims to generate noisy masks $\mathcal{M}_n$ at arbitrary timestep $t$ by adding Gaussian noise to encoded masks $\mathcal{M}_{enc}$. We first re-scale the scale of encoded masks from [0,1] to [-$b$,$b$], where $b$ is a scale factor. Afterwards, we add the Gaussian noise to encoded masks according to Eq. \ref{eq:diff}. The scale $1-\bar\alpha_t$ of added noise  depends on timestep $t$, where the large $t$ corresponds to  the large scale of noise. After adding noise, we clamp the noisy masks with the range of [-$b$, $b$].

\noindent\textbf{Loss formulation.} With the $N$ padded noisy masks, we can predict $N$  masks and corresponding classification scores. Afterwards, we perform bipartite matching between the predictions and ground-truths via Hungarian-based algorithm considering both mask and classification matching cost. After that, we can calculate the overall training losses as
\begin{equation}
L=\lambda_1 L_{cls}+ \lambda_2 L_{ce}+ \lambda_3 L_{dice},
\end{equation}
where $L_{cls}$ is cross entropy loss for classification, $L_{ce}$ and $L_{mask}$ are binary cross-entropy loss and dice loss for mask prediction. $\lambda_1,\lambda_2,\lambda_3$ indicate the loss weights, which are equal to 2.0, 5.0, 5.0.

\subsubsection{Inference}
Algorithm \ref{alg:sample} presents the inference details of our DFormer. We first generate noisy masks $\mathcal{M}_n$ with Gaussian distribution, and then predict instance masks with multiple sampling steps. In each sampling step, we first predict instance masks according to noisy masks at previous timestep, and then add noise to the predicted instance masks to generate  noisy masks at current timestep.

\begin{table*}[t!]
\centering
\footnotesize
\caption{\textbf{Panoptic segmentation comparison on MS-COCO \texttt{val2017} set} with 133 categories. Our DFormer outperforms the recent diffusion-based method Pix2Seq-$\mathcal{D}$ \cite{Chen_GenD_arXiv_2022} by 3.6\%.}
\begin{tabular}{r|c|c|c|ccc}
\toprule
Method & Backbone &\#Params & Epoch & PQ & PQ$^\text{Th}$ & PQ$^\text{St}$  \\
\midrule

Panoptic FPN \cite{Kirillov_PFPN_CPVR_2019}      & ResNet-50 & - &  12 & 39.0 & 45.9 & 28.7  \\
SOLOv2 \cite{Wang_SOLOv2_NIPS_2020}     & ResNet-50 & - & 36 & 42.1 & 49.6 &30.7 \\
DETR~\cite{Carion_DETR_ECCV_2020}       & ResNet-50 & - & 300 & 43.4 & 48.2 & 36.3  \\
Panoptic FCN   \cite{Li_PFCN_CPVR_2021}    & ResNet-50  & 37M & 12 & 43.6 & 49.3 &35.0 \\
MaX-DeepLab \cite{Wang_MaxDeepLab_CVPR_2021}       & MaX-S & 61.9M & 55 & 48.4 & 53.0 & 41.5  \\
MaskFormer~\cite{Cheng_MaskFormer_NeurIPS_2021} & ResNet-50 & 45M & 50 & 46.5 & 51.0 & 39.8  \\
K-Net~\cite{Zhang_KNet_NeurIPS_2021} & ResNet-50 & - & 36 & 47.1 & 51.7 & 40.3 \\
CMT-DeepLab~\cite{Yu_CMT_CVPR_2022} & ResNet-50 & - & 55  & 48.5 & - & - \\
Panoptic SegFormer~\cite{Li_PSegFormer_CVPR_2022} & ResNet-50 & 51M & 24 & 49.6 & 54.4 & 42.4 \\
kMaX-DeepLab~\cite{Yu_KMaX_ECCV_2022} & ResNet-50 & 57M & 50 & \textbf{53.0} & \textbf{58.3} & \textbf{44.9} \\
Mask2Former~\cite{Cheng_Mask2Former_CVPR_2022} & ResNet-50 & 44M & 50 & 51.9 & 57.7 & 43.0 \\
\midrule
\multicolumn{6}{l}{\textit{Diffusion-based approaches:\hfill}} \\
Pix2Seq-$\mathcal{D}$ (5-step) \cite{Chen_GenD_arXiv_2022} & ResNet-50  &  94.5M & 800& 47.5 & 52.2 & 40.3 \\
\textbf{DFormer (Ours)} & ResNet-50  &  44M & 50 &  \textbf{51.1} & \textbf{56.6}  & \textbf{42.8} \\

\bottomrule
\end{tabular}
\label{tab:sota_panoptic}
\vspace{-0.5cm}
\end{table*}
\section{Experiments}
\label{sec:exp}

\subsection{Datasets and metrics}
We adopt the large-scale dataset MS COCO \cite{Lin_COCO_ECCV_2014} for instance segmentation and panoptic segmentation, and the large-scale dataset ADE20K \cite{Zhou_ADE20K_CVPR_2017}  for semantic segmentation.

\noindent\textbf{MS COCO.} MS COCO \cite{Lin_COCO_ECCV_2014} is one of largest datasets for common vision tasks, such as object detection, instance segmentation, and panoptic segmentation. For object detection and instance segmentation, there are 80 object categories. For panoptic segmentation, there are 80 object (thing) categories and 53 stuff categories (\textit{i.e.,} totally 133 categories). It contains $\texttt{train2017}$, $\texttt{val2017}$, $\texttt{test-dev}$ sets, where the $\texttt{train2017}$ set has 118$k$ images, the $\texttt{val2017}$ set has 5$k$ images, and the $\texttt{test-dev}$ set has 20$k$ images. We adopt AP, AP$_{50}$, AP$_{75}$, AP$_{S}$, AP$_{M}$, AP$_{L}$ as evaluation metrics of instance segmentation \cite{Lin_COCO_ECCV_2014}, and use PQ, PQ$^\text{Th}$, PQ$^\text{St}$ as evaluation metrics of panoptic segmentation.

\noindent\textbf{ADE20k.} ADE20K \cite{Zhou_ADE20K_CVPR_2017} is one of largest datasets for semantic segmentation, which contains 150 semantic categories. The dataset contains $\texttt{train}$, $\texttt{val}$, $\texttt{test-dev}$ sets, where the $\texttt{train}$ set has 20$k$ images, the $\texttt{val}$ set has 2$k$ images, and the $\texttt{test}$ set has 3$k$ images. We adopt mIoU as evaluation metric of semantic segmentation.

\subsection{Implementation details}
We adopt the deep model ResNet50 \cite{He_ResNet_CVPR_2016} or Swin-T \cite{Liu_Swin_ICCV_2021} pre-trained on ImageNet as the backbone. The strides of feature maps $\mathcal{F}_p^{1,2,3,4}$ generated by pixel-level decoder are 32, 16, 8, and 4, and the number of feature channels is 256. In diffusion-based decoder, there are $L=9$ cascaded decoder layers. During training, each decoder layer is supervised with an auxiliary loss, and the noisy masks are padded to 100. During inference, we only keep the predicted masks  by last decoder layer. We adopt post-processing to obtain the expected output format for different segmentation tasks as in \cite{Cheng_MaskFormer_NeurIPS_2021,Cheng_Mask2Former_CVPR_2022}. The implementation details on different image segmentation tasks are summarized below.

\noindent\textbf{Instance and panoptic segmentation.} Our method is trained on four NVIDIA RTX3090 GPUs with the batch size of 12. During training, we adopt AdamW as the optimizer and step learning rate strategy, and there are 50 epochs totally. The initial learning is set as $1\times 10^{-4}$, and the weigh decay is set as 0.05. The learning rate is decreased at 0.9 and 0.95 fractions of total epochs by a factor of 10. We adopt large-scale jittering augmentation strategy as Mask2Former \cite{Cheng_Mask2Former_CVPR_2022}. During inference, we adopt the standard single-scale test, where the image is resized with the short edge  to 800 pixels and the longer edge up-to 1333 pixels.

\noindent\textbf{Semantic segmentation.} Our method is trained on four NVIDIA RTX3090 GPUs with the batch size of 16. During training, we adopt AdamW as the optimizer and poly learning rate schedule, and there are 160$k$ iterations totally. The initial learning is set as $1\times 10^{-4}$, and the weigh decay is set as $1\times 10^{-4}$. We employ random scale jitter for data augmentation as Mask2Former \cite{Cheng_Mask2Former_CVPR_2022} . During inference, we resize the images with the short edge  to 512 pixels.

\begin{table*}[t!]
\centering
\footnotesize
\caption{\textbf{Instance segmentation comparison on MS-COCO \texttt{val2017} and \texttt{test-dev} sets} with 80 categories. Our DFormer outperforms the recent diffusion-based method DiffusionInst \cite{Gu_DiffInst_arXiv_2022} with an absolute gain of 5.3\% on the \texttt{val2017}, and 6.2\% on the \texttt{test-dev}.}
\begin{tabular}{r|c|c|c|cc|ccc}
\toprule
Method & Backbone & Epoch & AP & AP$_{50}$ & AP$_{75}$ & AP$_{S}$ & AP$_{M}$ & AP$_{L}$ \\
\midrule
\multicolumn{8}{c}{\textbf{Comparison on \texttt{val2017} set}\hfill} \\
Mask R-CNN \cite{He_MaskRCNN_ICCV_R2017} & ResNet-50 & 36  & 37.1 & 58.5 & 39.7 & 18.7 & 39.6 & 53.9 \\
MaskFormer \cite{Cheng_MaskFormer_NeurIPS_2021} & ResNet-50  &  300 & 34.0 & -& -& 16.4 & 37.8 & 54.2\\
SOLO \cite{Wang_SOLO_ECCV_2020} & ResNet-50  & 36 & 35.8 & 56.7 & 37.9 & 14.3 & 39.3 & 53.2\\
SOLOv2 \cite{Wang_SOLOv2_NIPS_2020} & ResNet-50  & 36 & 37.8 & 58.5& 40.4& - & - & -\\
CondInst \cite{Tian_CondInst_ECCV_2020} & ResNet-50  & 36 & 37.5 & 58.5& 40.1& 18.7 & 41.0 & 53.3\\
DynaMask \cite{Li_DynaMask_CVPR_2023} & ResNet-50  & 36 & 38.2 & 58.1 & \textbf{41.5} & 20.5 & 40.8 & 52.7\\ 
K-Net \cite{Zhang_KNet_NeurIPS_2021} & ResNet-50  & 36 & 37.8 & \textbf{60.3}& 39.9& 16.9 & 41.2 & 57.5\\
Mask2Former \cite{Cheng_Mask2Former_CVPR_2022} & ResNet-50  & 50 & \textbf{43.7} & -& -& \textbf{23.4} & \textbf{47.2} & \textbf{64.8}\\
\midrule
\multicolumn{6}{l}{\textit{Diffusion-based approaches:\hfill}} \\
DiffusionInst \cite{Gu_DiffInst_arXiv_2022} & ResNet-50 & 60  & 37.3 & 60.3 & 39.3 & 18.9 & 40.1 & 54.7\\
\textbf{DFormer (Ours)} & ResNet-50  & 50 & \textbf{42.6}  & \textbf{64.8}  & \textbf{45.8} & \textbf{22.3} & \textbf{45.7} & \textbf{64.2}\\
\midrule
\multicolumn{8}{c}{\textbf{Comparison on \texttt{test-dev} set}\hfill} \\
Mask R-CNN \cite{He_MaskRCNN_ICCV_R2017} &  ResNet-50 & 36 & 37.4 & 59.5 & 40.1 & 18.6 & 39.8 & 51.6\\ 
CondInst \cite{Tian_CondInst_ECCV_2020} & ResNet-50  & 36 & 38.8 & 60.4 & 41.5& 21.1 &41.1 &51.0 \\
SOLO \cite{Wang_SOLO_ECCV_2020} & ResNet-50  &  72 & 36.8 & 58.6 & 39.0 & 15.9 & 39.5 & 52.1 \\ 
SOLOv2 \cite{Wang_SOLOv2_NIPS_2020} & ResNet-50  &  36 & 38.8 & 59.9 & 41.7 & 16.5 & 41.7 & 56.2 \\ 
RefineMask \cite{Zhan_RefineMask_CVPR_2021} & ResNet-50  & 36 & \textbf{40.2} & - & - & \textbf{27.5} & \textbf{50.6} & \textbf{60.7}\\ 
QueryInst \cite{Fang_QueryInst_ICCV_2021} & ResNet-50  & 36 & 40.1 & \textbf{62.3} &  \textbf{43.4} &  23.3 & 42.1 & 52.0\\ 
K-Net \cite{Zhang_KNet_NeurIPS_2021} & ResNet-50  & 36 & 38.4 & 61.2 & 40.9 & 17.4 & 40.7 & 56.2 \\ 
SparseInst \cite{Cheng_SparseInst_CVPR_2022} & ResNet-50  & 36 & 37.9 & 59.2 & 40.2 & 15.7 & 39.4 & 56.9\\ 
SOLQ \cite{Dong_SOLQ_NeurIPS_2021} & ResNet-50  & 50 & 39.7 & - & - & 21.5 &42.5& 53.1\\ 
Mask Transfiner \cite{Ke_MTrans_CVPR_2022} & ResNet-50  & 36 & 39.4 &  - &  - &  22.3 & 41.2 & 50.2\\ 

\midrule
\multicolumn{6}{l}{\textit{Diffusion-based approaches:\hfill}} \\
DiffusionInst \cite{Gu_DiffInst_arXiv_2022} & ResNet-50   & 60 & 37.1 & - & - & 19.4 & 39.7 &49.3\\
\textbf{DFormer (Ours)} & ResNet-50   & 50 & \textbf{43.3}  & \textbf{65.8}   & \textbf{46.7}  & \textbf{15.6} & \textbf{61.5} & \textbf{78.1}\\
\bottomrule
\end{tabular}
\label{tab:sota_instance}
\vspace{-0.5cm}
\end{table*}

\subsection{State-of-the-art comparison}
\noindent\textbf{Panoptic segmentation.} We compare our DFormer with state-of-the-art methods for panoptic segmentation on MS COCO \texttt{val2017} set in Table \ref{tab:sota_panoptic}. All the methods use ResNet-50 as the backbone. Compared to the recent diffusion-based method Pix2Seq-$\mathcal{D}$ \cite{Chen_GenD_arXiv_2022}, our DFormer achieves significant improvement while being efficient in terms of parameters. DFormer outperforms Pix2Seq-$\mathcal{D}$ with an absolute gain of 3.6\% on PQ, 4.4\% on PQ$^{Th}$, and 2.5\% on PQ$^{St}$. In addition, our DFormer has fewer parameters and training epochs. Compared to other state-of-the-art methods, our DFormer  achieves a competitive performance. 

\noindent\textbf{Instance segmentation.} We compare our DFormer with several state-of-the-art methods for instance segmentation on MS COCO \texttt{val2017} and \texttt{test-dev} sets in Table \ref{tab:sota_instance}. All the methods use ResNet-50 as the backbone. Compared to the recent diffusion-based method DiffusionInst \cite{Gu_DiffInst_arXiv_2022}, our DFormer achieves significant improvement in performance. In terms of overall AP, our DFormer outperforms DiffusionInst by an absolute gain of 5.3\% on the \texttt{val2017} set and 6.2\% on the \texttt{test-dev} set.  Compared to other state-of-the-art methods, our DFormer   achieves a competitive performance. For instance, our DFormer outperforms Mask Transfiner \cite{Ke_MTrans_CVPR_2022} by 3.9\% on the \texttt{test-dev} set.

\begin{table*}[t!]
\centering
\footnotesize
\caption{\textbf{Semantic segmentation comparison on ADE20k \texttt{val} set} with 155 categories. Our method are also competitive with state-of-the-art methods.}
\begin{tabular}{r|c|c}
\toprule
Method & Backbone  & mIoU \\
\midrule

UperNet \cite{Xiao_UperNet_ECCV_2018} & Swin-T  & 44.5\\

K-Net \cite{Zhang_KNet_NeurIPS_2021} & ResNet-50   & 44.6 \\ 
K-Net \cite{Zhang_KNet_NeurIPS_2021} & Swin-T  & 45.8 \\ 
kMaX-Deeplab \cite{Yu_KMaX_ECCV_2022} & ResNet-50 & 45.0 \\
MaskFormer \cite{Cheng_MaskFormer_NeurIPS_2021} & ResNet-50  & 44.5 \\
MaskFormer \cite{Cheng_MaskFormer_NeurIPS_2021} & Swin-T & 46.7 \\
Mask2Former \cite{Cheng_Mask2Former_CVPR_2022} & ResNet-50  & 47.2 \\
Mask2Former \cite{Cheng_Mask2Former_CVPR_2022} & Swin-T& \textbf{47.7} \\
\midrule
\multicolumn{3}{l}{\textit{Diffusion-based approaches:\hfill}} \\
DDP \cite{Ji_DDP_arXiv_2023} & Swin-T  &  46.1 \\
\textbf{DFormer (Ours)}  & ResNet-50 &  46.7 \\
\textbf{DFormer (Ours)}  & Swin-T  & \textbf{48.3}  \\
\bottomrule
\end{tabular}
\vspace*{-0.5cm}
\label{tab:sota_semantic}
\end{table*}

\noindent\textbf{Semantic segmentation.} Lastly, we compare our DFormer with several state-of-the-art methods on ADE20K \texttt{val} set \cite{Zhou_ADE20K_CVPR_2017} in Table \ref{tab:sota_semantic}. We present the results using the backbone ResNet50 and Swin-T. When using the backbone Swin-T, our DFormer outperforms the recent diffusion-based method DDP \cite{Ji_DDP_arXiv_2023} with an absolute gain of 2.2\%. Moreover, our DFormer  performs favorably against existing methods in literature. Compared to K-Net \cite{Zhang_KNet_NeurIPS_2021}, DFormer achieves an absolute gain of 2.5\% using the same backbone. Further, DFormer also obtains improved performance with AP of 48.3, compared to the recent Mask2Former \cite{Cheng_Mask2Former_CVPR_2022}, when using the same Swin-T backbone. 

\begin{table}[t]
\centering
\footnotesize
\caption{\textbf{Ablation study.} We perform experiments  on MS COCO \texttt{val2017} for instance segmentation  using  ResNet50 except (c). In (c), we show the results of different models on different tasks.}
\subfloat[Number of decoder layers\label{tab:numlayer}]{
\begin{tabular}{r|c|cc}
 & AP & AP$_{50}$ & AP$_{75}$\\
\midrule
3 & 41.1 & 62.4  & 44.3  \\
6 & 42.4 & 64.4 & 45.7  \\
9 & 42.6 & 64.8 & 45.8  \\
\end{tabular}}\hspace{1mm}
\subfloat[Different diffusion steps\label{tab:steps}]{
\begin{tabular}{r|ccc}
 & AP & AP$_{50}$ & AP$_{75}$\\
\midrule
1 & 42.6 & 64.8  & 45.8 \\
2 & 42.7 & 64.8 & 45.9  \\
4 & 42.7 & 64.8 & 45.9 \\
\end{tabular}}\hspace{1mm}
\subfloat[Results on ResNet-50 and Swin-T\label{tab:swin-t}]{
\begin{tabular}{c|cc}
&  ResNet-50 & Swin-T\\
\midrule
Instance & 42.6& 44.4 \\
Semantic  &46.7 &48.3\\
Panoptic  & 51.1 &52.5 \\
\end{tabular}}\par
\subfloat[Number of  noisy masks\label{tab:ablation:nummask}]{
\begin{tabular}{r|ccc}
 & AP & AP$_{50}$ & AP$_{75}$\\
\midrule
50 & 40.7 & 62  & 43.4  \\
100 & 42.6 & 64.8 & 45.8  \\
150 & 42.7 & 64.9 & 46.1  \\
200 & 42.5 & 64.4 & 45.7  \\
\end{tabular}}\hspace{3mm}
\subfloat[Mask encoding strategy and scale factor\label{tab:noisymasks}]{
\begin{tabular}{r|ccc}
 & AP & AP$_{50}$ & AP$_{75}$\\
\midrule
Binary with $b$=1 & 41.2 & 63.4  & 44.2 \\
Binary with $b$=0.1 & 42.6 & 64.8 & 45.8  \\
Random with $b$=1 & 42.3 & 64.3 &45.4   \\
Random with $b$=0.1 & 42.5 & 64.6 & 45.5  \\
\end{tabular}}
\label{tab:ablation:main}
\vspace{-1cm}
\end{table}

\subsection{Ablation study}

\noindent\textbf{Decoder layers.} Table \ref{tab:numlayer} presents the impact of different number of decoder layers. It achieves a mask AP score of 41.1 when the number is equal to 3,
whereas it obtains a mask AP score of 42.6 when the number is equal to 9. We therefore adopt 9 decoder layers in our final model.

\begin{figure}[t]
\centering
\subfloat[Panoptic segmentation]
{\label{fig:a}\includegraphics[width=0.95\linewidth]{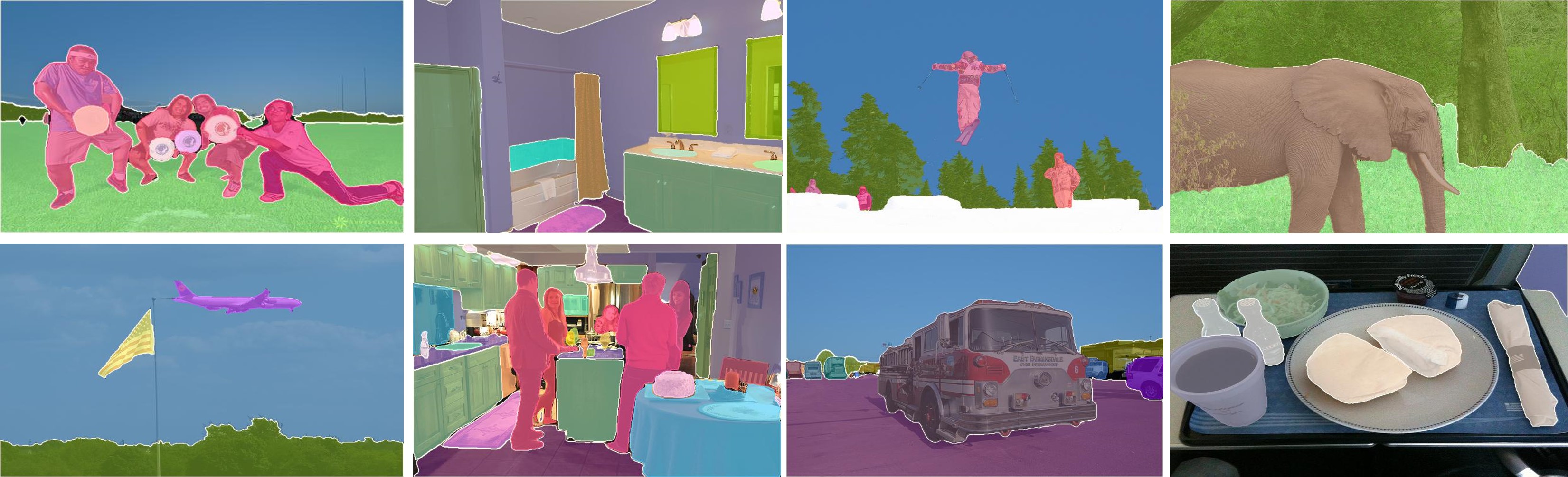}}\\[-0.2ex]
\subfloat[Instance segmentation]{\label{fig:b}\includegraphics[width=0.95\linewidth]{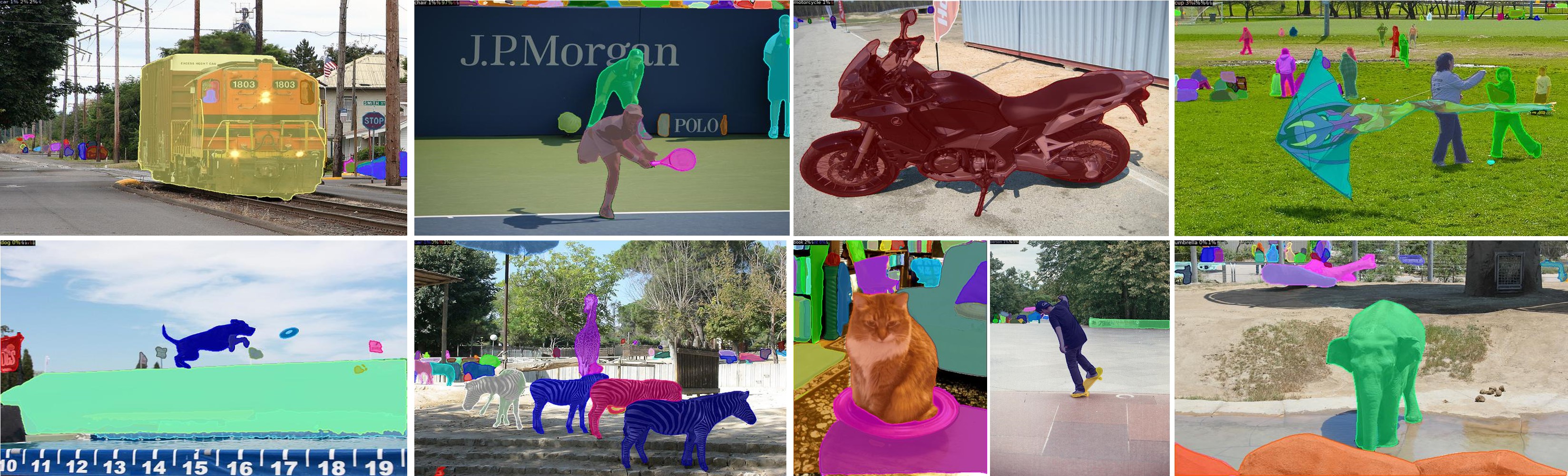}}\\[-0.2ex]
\subfloat[Semantic segmentation]{\label{fig:c}\includegraphics[width=0.95\linewidth]{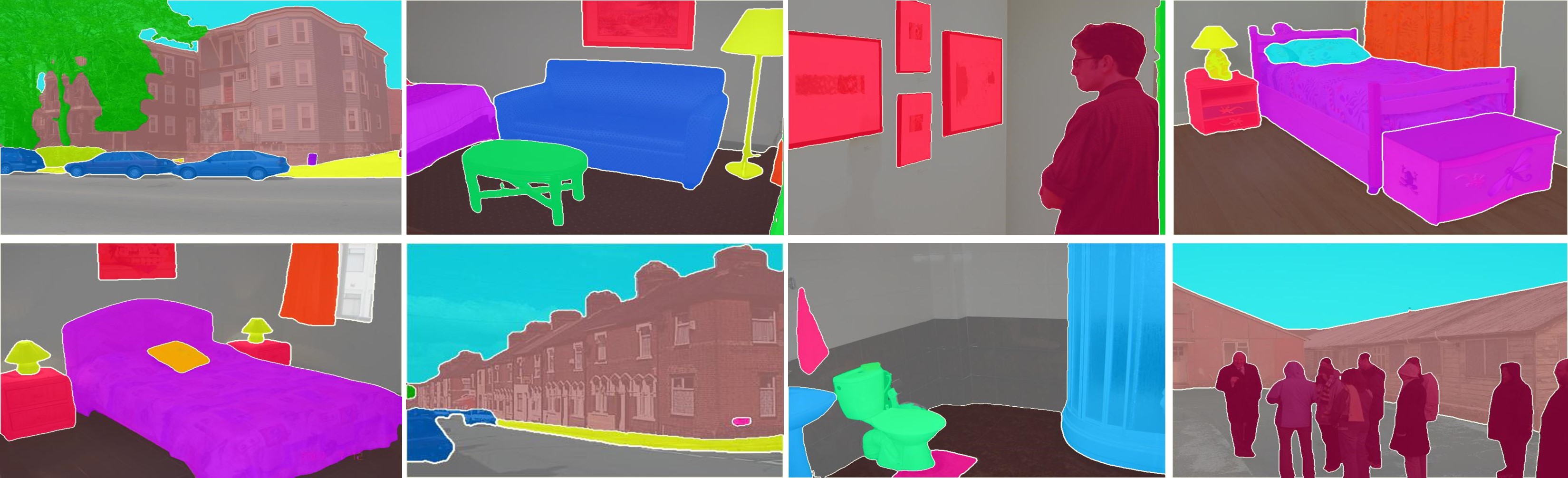}}\\[-0.7ex]
\caption{\textbf{Qualitative results  on three image segmentation tasks}.  DFormer achieves  high-quality segmentation, with the same architecture, on three tasks under different challenging scenarios.}
\vspace{-0.5cm}
\label{fig:vis}
\end{figure}

\noindent\textbf{Timesteps during inference.} Table \ref{tab:steps}  presents the impact of different timesteps during inference. It almost achieves the best performance using 1-step forward process. The multi-step forward process  provides a  slight improvement in performance.

\noindent\textbf{Different backbones on different tasks.} Table \ref{tab:swin-t} shows the impact of different models on different tasks. Swin-T achieves better performance than ResNet-50 on all three segmentation tasks.

\noindent\textbf{Number of noisy masks.} Table \ref{tab:ablation:nummask} shows the impact of different numbers of noisy masks during inference. The default setting is using 100 which is same as the training. We observe that a larger number of noisy masks does not affect the performance.

\noindent\textbf{Mask encoding and scale strategies.} Table \ref{tab:noisymasks} shows the impact of different mask encoding and scale strategies. As discussed earlier, we introduce two different mask encoding strategies, including binary and random shuffling-based strategy. When mask scale $b$ is equal to 1, the random shuffling-based encoding  achieves a better performance than its binary counterpart. When the mask scale $b$ is equal to 0.1, the random shuffling strategy  is slightly inferior than the binary one. Therefore, we adopt the binary strategy with the scale as 0.1.

\noindent\textbf{Qualitative results.} Fig. \ref{fig:vis} presents example qualitative results of our DFormer on the three image segmentation tasks.  Our DFormer provides  high-quality segmentation results on these different tasks under the challenging scenarios, such as segmenting occluded objects (\textit{e.g.,} row 2, column 2) and small-scale objects (\textit{e.g.,} row 3, column 4). Additional results are presented in the suppl. material.

\section{Conclusion}
\label{sec:conclusion}

We propose a diffusion-guided transformer for universal image segmentation. Our DFormer views universal image segmentation as diffusion-based mask classification. During training, we first perform diffusion process by adding Gaussian noise to ground-truth masks, and then learn a transformer-guided model to predict ground-truth mask from noisy masks. At inference, we directly predict masks and corresponding categories from Gaussian noise. Experiments are performed on panoptic, instance and semantic segmentation tasks to validate the effectiveness of our DFormer. Dformer significantly outperforms existing diffusion-based methods on all three image segmentation tasks. \\
\noindent\textbf{Limitations and future directions.} We observe that our proposed diffusion-guided method is slightly inferior to state-of-the-art methods on small objects on instance segmentation. A potential future direction is to explore improving performance on small objects with diffusion model.

\appendix
\section{Visualization of diffusion process}
\begin{figure}[ht]
\centering
\includegraphics[width=0.95\linewidth]{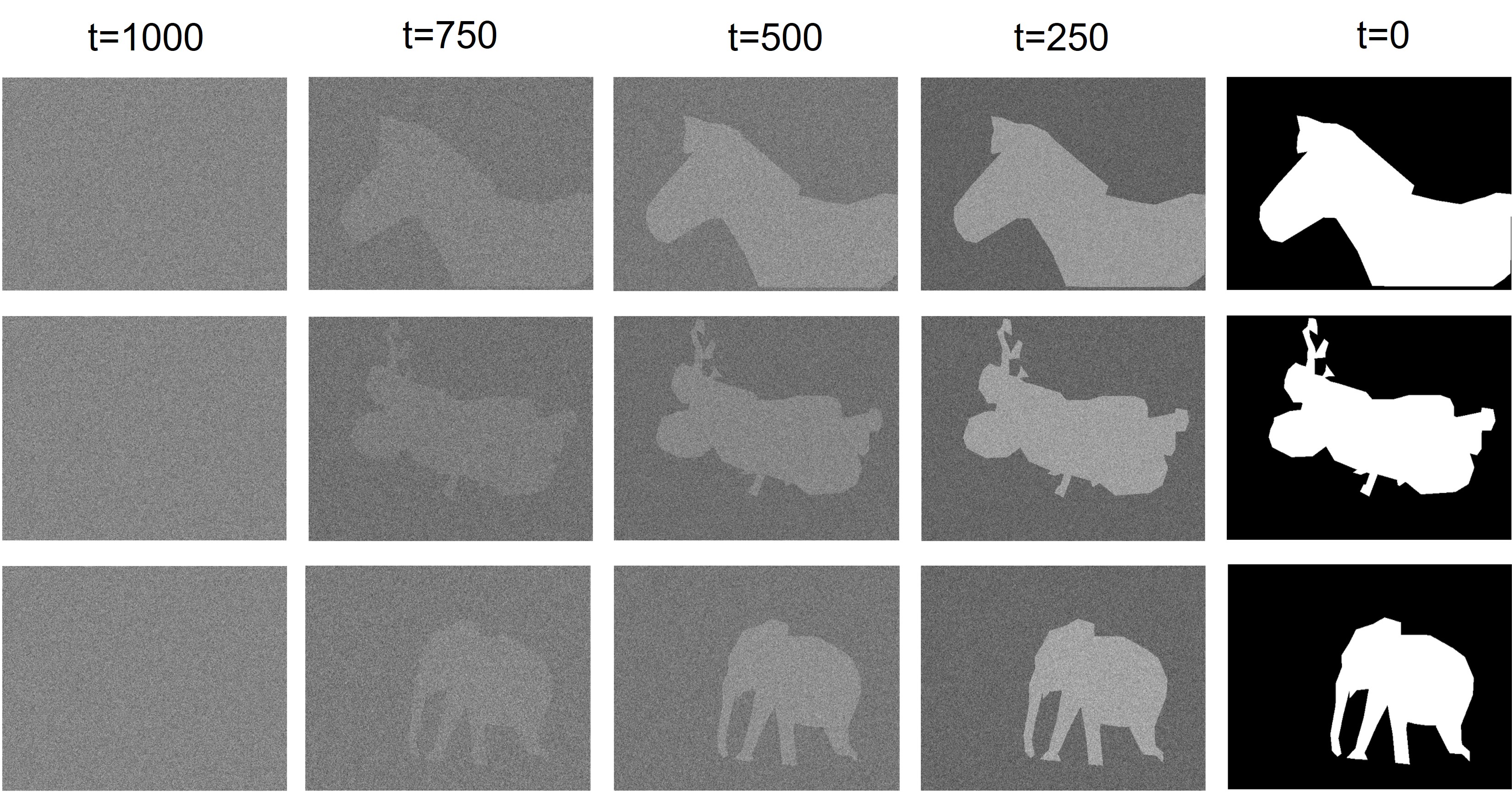}
\caption{Visualization of diffusion process}
\label{fig:noisymask}
\end{figure}

Fig. \ref{fig:noisymask}  shows visualization of diffusion process by adding Gaussian noise to the ground-truth mask. The total timestep $T$ is equal to 1000. When timestep $t$ is equal to $T$, the noisy mask will be Gaussian noise. With the decrease of $t$, the noisy mask become more clear. When timestep $t$ is equal to $T$, the noisy mask will be the ground-truth mask. Our DFormer presents high-quality segmentation results on different tasks, using a single architecture.

\section{Extensions on video instance segmentation}
We extend DFormer to video instance segmentation by adding a tracking branch. Table \ref{tab:vis} shows the results on video instance segmentation. DFormer has also demonstrated competitive performance on video instance segmentation.
\vspace{-3mm}
\begin{table*}[ht]
\centering
\footnotesize
\caption{\textbf{Video instance segmentation on YouTube-VIS 2019 \texttt{val} set}}
\begin{tabular}{r|cccccc}
\toprule
Method & Backbone  &AP&AP$_{50}$ & AP$_{75}$&AR$_{1}$&AR$_{10}$\\
\midrule
SeqFormer \cite{Wu_SeqFormer_ECCV_2022} &R50&45.1&66.9&50.5&45.6&54.6\\
Mask2Former-VIS \cite{Cheng_M2FVIS_arXiv_2021} &R50&46.4&68.0&50.0&-&-\\
MinVIS \cite{Huang_MinVIS_NeurIPS_2022} &R50&47.4&69&52.1&45.7&55.7\\
\textbf{DFormer (Ours)}&R50&46.5&69.7&49.2&44.8&54.3\\
\bottomrule
\end{tabular}
\label{tab:vis}
\vspace{-0.5cm}
\end{table*}
\section{More visualization of segmentation results}
\begin{figure}[ht]
\centering
\includegraphics[width=0.95\linewidth]{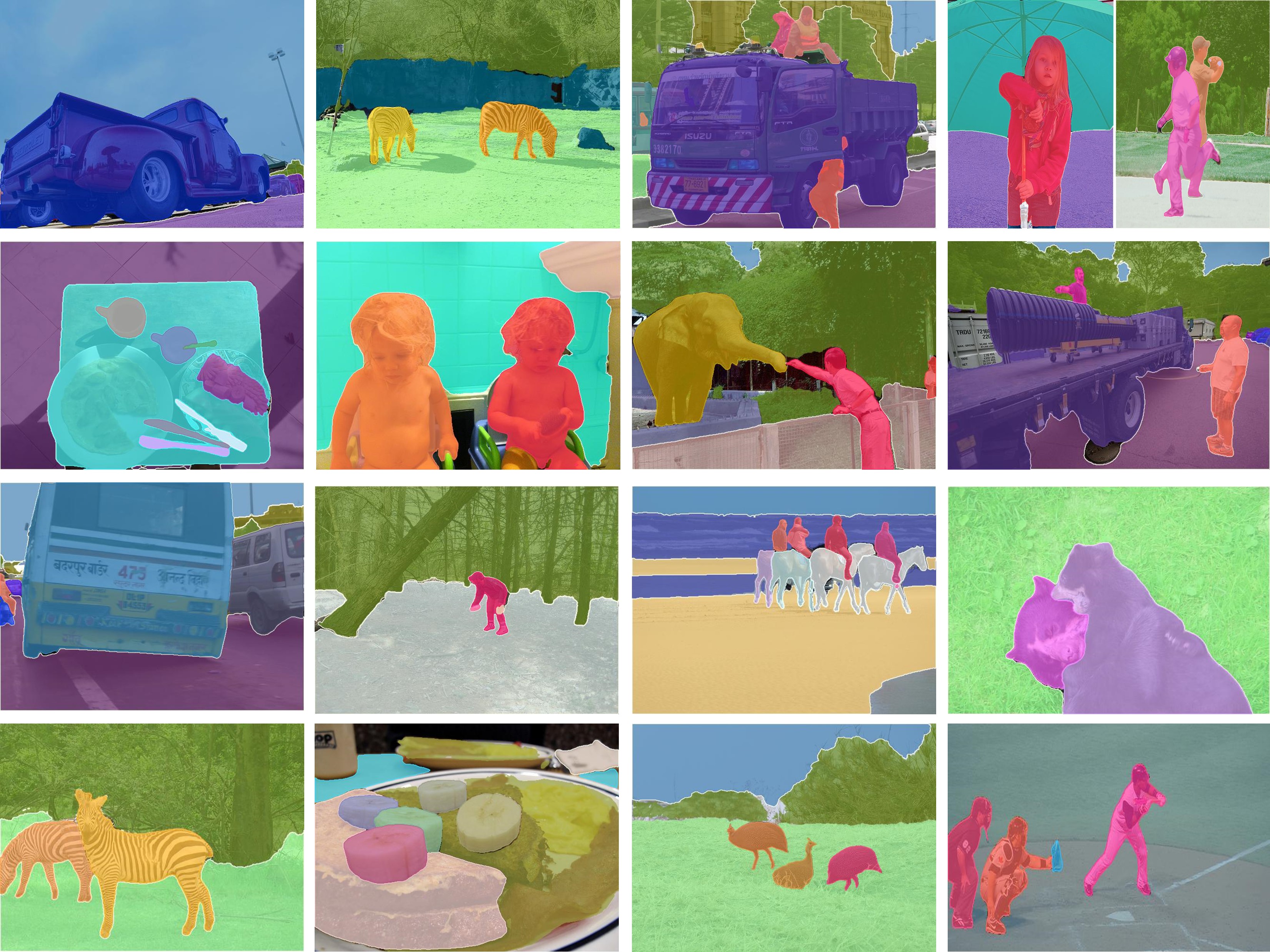}
\caption{Visualization of panoptic segmentation on COCO2017 val set.}
\label{fig:bupac}
\end{figure}

\begin{figure}[h!]
\centering
\includegraphics[width=0.95\linewidth]{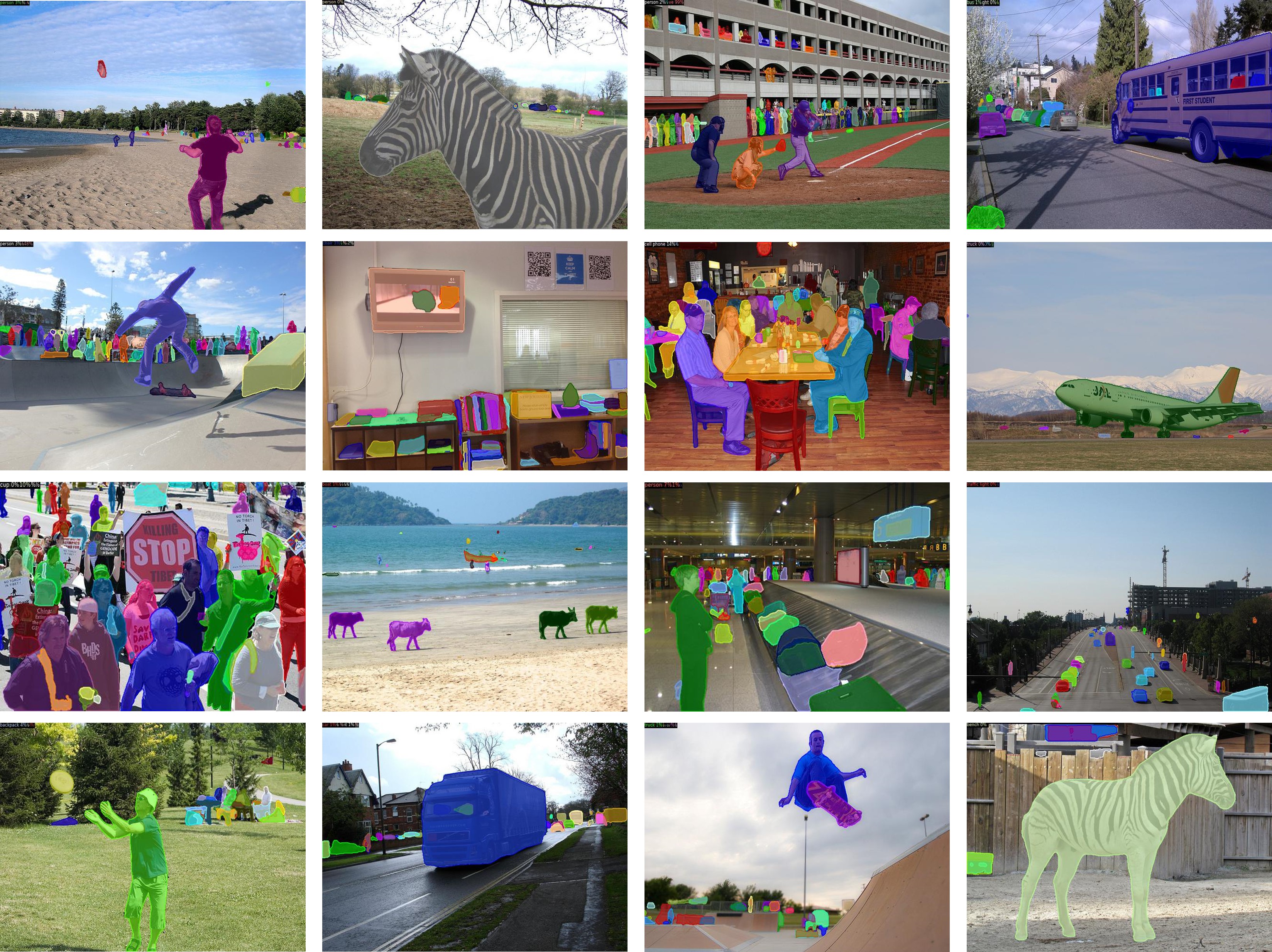}
\caption{ Visualization of instance segmentation on COCO2017 val set.}
\label{fig:buins}
\end{figure}

\begin{figure}[ht]
\centering
\includegraphics[width=0.92\linewidth]{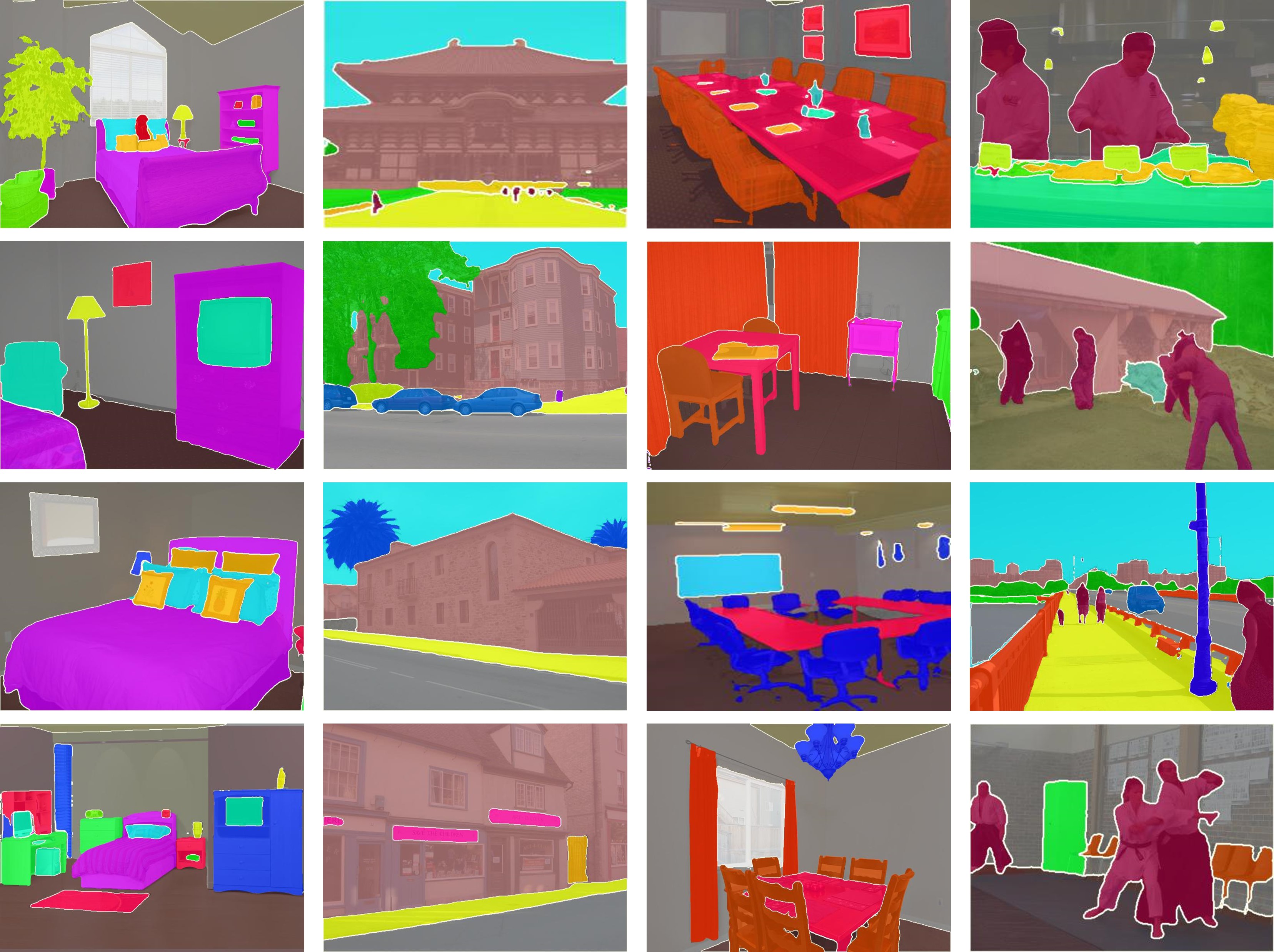}
\caption{Visualization of semantic segmentation on ADE20K val set.}
\label{fig:busem}
\end{figure}

Fig. \ref{fig:bupac} shows more results of panoptic segmentation results of our DFormer, Fig. \ref{fig:buins} shows more results of instance segmentation results of our DFormer, and Fig. \ref{fig:busem} shows more results of semantic segmentation results of our DFormer.

\vspace{9mm}
\vspace{-5mm}
\vspace{7mm}

\medskip
{\small
\bibliographystyle{unsrt}
\bibliography{egbib}
}

\end{document}